# *Module Theorem for The General Theory of Stable Models*

Joseph Babb and Joohyung Lee
*School of Computing, Informatics, and Decision Systems Engineering*
*Arizona State University, Tempe, AZ, USA*
(*e-mail:* {Joseph.Babb, joolee}@asu.edu)

**Abstract**

The module theorem by Janhunen *et al.* demonstrates how to provide a modular structure in answer set programming, where each module has a well-defined input/output interface which can be used to establish the compositionality of answer sets. The theorem is useful in the analysis of answer set programs, and is a basis of incremental grounding and reactive answer set programming. We extend the module theorem to the general theory of stable models by Ferraris *et al.* The generalization applies to non-ground logic programs allowing useful constructs in answer set programming, such as choice rules, the count aggregate, and nested expressions. Our extension is based on relating the module theorem to the symmetric splitting theorem by Ferraris *et al.* Based on this result, we reformulate and extend the theory of incremental answer set computation to a more general class of programs.

## 1 Introduction

The module theorem (Oikarinen and Janhunen 2008; Janhunen et al. 2009) demonstrates how to provide a modular structure for logic programs under the stable model semantics, where each module has a well-defined input/output interface which can be used to establish the compositionality of answer sets of different modules. The theorem was shown to be useful in the analysis of answer set programs and was used as a basis of incremental grounding (Gebser et al. 2008) and reactive answer set programming (Gebser et al. 2011), resulting in systems ICLINGO and OCLINGO.

The module theorem was stated for normal logic programs and SMODELS programs in (Oikarinen and Janhunen 2008) and for disjunctive logic programs in (Janhunen et al. 2009), but both papers considered ground programs only. In this paper we extend the module theorem to non-ground programs, or more generally, to first-order formulas under the stable model semantics proposed by Ferraris et al. (2011). We derive the generalization by relating the module theorem to the symmetric splitting theorem by Ferraris et al. (2009). This is expected in some sense as the symmetric splitting theorem looks close to the module theorem and is already applicable to first-order formulas under the stable model semantics (Ferraris et al. 2011). Since non-ground logic programs can be understood as a special class of first-order formulas under the stable model semantics, the theorem can be applied to split these programs. In addition, as the semantics of choice rules and the count aggregate in answer set programming is understood as shorthand for some first-order formulas (Lee et al. 2008), the splitting theorem can also be applied to non-ground programs containing such constructs.

The precise relationship between the module theorem and the splitting theorem has not been established, partly because there is some technical gap that needs to be closed. While the splitting theorem is applicable to more general classes of programs in most cases, there are some cases where the module theorem allows us to split, but the splitting theorem does not.

In order to handle this issue, we first extend the splitting theorem to allow this kind of generality. We then add modular structures to the splitting theorem, and provide a mechanism of composing partial interpretations for each module. This new theorem serves as the module theorem for the general theory of stable models.

The paper is organized as follows. In the next section, we review the stable model semantics from (Ferraris et al. 2011), the splitting theorem, and the module theorem. In Section 3 we provide a generalization of the splitting theorem, which closes the gap between the module theorem and the splitting theorem. In Section 4 we present the module theorem for the general theory of stable models, which extends both the previous splitting theorem and the previous module theorem. We give an example of the generalized module theorem in Section 5 and show how it serves as a foundation for extending the theory of incremental answer set computation in Section 6.

## 2 Preliminaries

### *2.1 Review: General Theory of Stable Models*

This review follows the definition by Ferraris et al. (2011). There, stable models are defined in terms of the SM operator, which is similar to the circumscription operator CIRC (Lifschitz 1994).

Let $\mathbf{p}$ be a list of distinct predicate constants $p_1, \dots, p_n$, and let $\mathbf{u}$ be a list of distinct predicate variables $u_1, \dots, u_n$. By $\mathbf{u} \leq \mathbf{p}$ we denote the conjunction of the formulas $\forall \mathbf{x}(u_i(\mathbf{x}) \rightarrow p_i(\mathbf{x}))$ for all $i = 1, \dots, n$, where $\mathbf{x}$ is a list of distinct object variables whose length is the same as the arity of $p_i$. Expression $\mathbf{u} < \mathbf{p}$ stands for $(\mathbf{u} \leq \mathbf{p}) \wedge \neg(\mathbf{p} \leq \mathbf{u})$. For instance, if $p$ and $q$ are unary predicate constants then $(u, v) < (p, q)$ is

$$\forall x(u(x) \rightarrow p(x)) \wedge \forall x(v(x) \rightarrow q(x)) \wedge \neg\Big(\forall x(p(x) \rightarrow u(x)) \wedge \forall x(q(x) \rightarrow v(x))\Big).$$

For any first-order formula $F$, $\mathrm{SM}[F;\mathbf{p}]$ is defined as

$$F \wedge \neg\exists\mathbf{u}((\mathbf{u} < \mathbf{p}) \wedge F^*(\mathbf{u})),$$

where $F^*(\mathbf{u})$ is defined recursively as follows: [1]

- $p_i(\mathbf{t})^* = u_i(\mathbf{t})$ for any list $\mathbf{t}$ of terms;
- $F^* = F$ for any atomic formula $F$ (including $\bot$ and equality) that does not contain members of $\mathbf{p}$;
- $(F \wedge G)^* = F^* \wedge G^*$;
- $(F \vee G)^* = F^* \vee G^*$;
- $(F \rightarrow G)^* = (F^* \rightarrow G^*) \wedge (F \rightarrow G)$;
- $(\forall x F)^* = \forall x F^*$;
- $(\exists x F)^* = \exists x F^*$.

When $F$ is a sentence, the models of $\mathrm{SM}[F;\mathbf{p}]$ are called the $\mathbf{p}$-*stable* models of $F$. Intuitively, they are the models of $F$ that are "stable" on $\mathbf{p}$. We will often simply write $\mathrm{SM}[F]$ in place of $\mathrm{SM}[F;\mathbf{p}]$ when $\mathbf{p}$ is the list of all predicate constants occurring in $F$, and often identify $\mathbf{p}$ with the corresponding set if there is no confusion.

By an *answer set* of $F$ that contains at least one object constant we understand an Herbrand interpretation of $\sigma(F)$ that satisfies $\mathrm{SM}[F]$, where $\sigma(F)$ is the signature consisting of the object, function and predicate constants occurring in $F$.

The answer sets of a logic program $\Pi$ are defined as the answer sets of the FOL-representation of $\Pi$ (i.e., the conjunction of the universal closures of implications corresponding to the rules). For example, the FOL-representation $F$ of the program

$$\begin{array}{l} p(a) \\ q(b) \\ r(x) \leftarrow p(x), not\ q(x) \end{array}$$

is

$$p(a) \wedge q(b) \wedge \forall x(p(x) \wedge \neg q(x) \rightarrow r(x)) \tag{1}$$

and $\mathrm{SM}[F]$ is

$$\begin{array}{l} p(a) \wedge q(b) \wedge \forall x(p(x) \wedge \neg q(x) \rightarrow r(x)) \\ \quad \wedge\neg\exists uvw(((u,v,w) < (p,q,r)) \wedge u(a) \wedge v(b) \\ \qquad\qquad \wedge\forall x((u(x) \wedge (\neg v(x) \wedge \neg q(x)) \rightarrow w(x)) \wedge (p(x) \wedge \neg q(x) \rightarrow r(x)))), \end{array}$$

which is equivalent to the first-order sentence

$$\forall x(p(x) \leftrightarrow x = a) \wedge \forall x(q(x) \leftrightarrow x = b) \wedge \forall x(r(x) \leftrightarrow (p(x) \wedge \neg q(x))) \tag{2}$$

(Ferraris et al. 2007, Example 3). The stable models of $F$ are any first-order models of (2). The only answer set of $F$ is the Herbrand model $\{p(a),\ q(b),\ r(a)\}$.

Ferraris et al. (2011) show that this definition of an answer set, when applied to the syntax of logic programs, is equivalent to the traditional definition of an answer set that is based on grounding and fixpoints (Gelfond and Lifschitz 1988).

[1] We understand $\neg F$ as shorthand for $F \rightarrow \bot$.

### *2.2 Review: Symmetric Splitting Theorem*

We say that an occurrence of a predicate constant, or any other subexpression, in a formula $F$ is *positive* if the number of implications containing that occurrence in the antecedent is even (recall that we treat $\neg G$ as shorthand for $G \rightarrow \bot$). We say that the occurrence is *strictly positive* if the number of implications in $F$ containing that occurrence in the antecedent is 0. For example, in (1), both occurrences of $q$ are positive, but only the first one is strictly positive. A *rule* of $F$ is an implication that occurs strictly positively in $F$.

A formula $F$ is called *negative* on a list $\mathbf{p}$ of predicate constants if members of $\mathbf{p}$ have no strictly positive occurrences in $F$. For example, formula (1) is negative on $\{s\}$, but is not negative on $\{p, q\}$. A formula of the form $\neg F$ (shorthand for $F \rightarrow \bot$) is negative on any list of predicate constants.

The following definition of a dependency graph is from (Lee and Palla 2012), which is similar to the one from (Ferraris et al. 2009), but may contain less edges.

*Definition 1* (*Predicate Dependency Graph*)
The *predicate dependency graph* of a first-order formula $F$ relative to $\mathbf{p}$, denoted by $\mathrm{DG}[F;\mathbf{p}]$, is the directed graph that

- has all members of $\mathbf{p}$ as its vertices, and
- has an edge from $p$ to $q$ if, for some rule $G \rightarrow H$ of $F$,
  - $p$ has a strictly positive occurrence in $H$, and
  - $q$ has a positive occurrence in $G$ that does not belong to any subformula of $G$ that is negative on $\mathbf{p}$.

For example, $\mathrm{DG}[(1);\ p, q, r]$ has the vertices $p, q$, and $r$, and a single edge from $r$ to $p$.

*Theorem 1* (*Splitting Theorem, (Ferraris et al. 2009)*)
Let $F$, $G$ be first-order sentences, and let $\mathbf{p}$, $\mathbf{q}$ be finite disjoint lists of distinct predicate constants. If

(a) each strongly connected component of the predicate dependency graph of $F \wedge G$ relative to $\mathbf{p}$, $\mathbf{q}$ is a subset of $\mathbf{p}$ or a subset of $\mathbf{q}$,
(b) $F$ is negative on $\mathbf{q}$, and
(c) $G$ is negative on $\mathbf{p}$

then

$$\mathrm{SM}[F \wedge G;\ \mathbf{pq}] \leftrightarrow \mathrm{SM}[F;\ \mathbf{p}] \wedge \mathrm{SM}[G;\ \mathbf{q}]$$

is logically valid.

Theorem 1 is slightly more generally applicable than the version of the splitting theorem from (Ferraris et al. 2009) as it refers to the refined definition of a dependency graph above instead of the one considered in (Ferraris et al. 2009).

*Example 1*
Theorem 1 tells us that $\mathrm{SM}[(1)]$ is equivalent to

$$\mathrm{SM}[p(a) \wedge q(b);\ p, q] \wedge \mathrm{SM}[\forall x(p(x) \wedge \neg q(x) \rightarrow r(x));\ r].$$

### *2.3 Review: DLP-Modules and Module Theorem*

Janhunen et al. (2009) considered rules of the form

$$a_1; \dots; a_n \leftarrow b_1, \dots, b_m, \imath not\ c_1, \dots, \imath not\ c_k \tag{3}$$

where $n, m, k \geq 0$ and $a_1, \dots, a_n, b_1, \dots b_m, c_1, \dots, c_k$ are propositional atoms. They define a *DLP-module* as a quadruple $(\Pi, \mathcal{I}, \O, \mathcal{H})$, where $\Pi$ is a finite propositional disjunctive logic program consisting of rules of the form (3), and $\mathcal{I}$, $\O$, and $\mathcal{H}$ are finite sets of propositional atoms denoting the input, output, and hidden atoms, respectively, such that (*i*) the sets of input, output, and hidden atoms are disjoint; (*ii*) every atom occurring in $\Pi$ is either an input, output, or hidden atom; (*iii*) every rule in $\Pi$ with a nonempty head contains at least one output or hidden atom.

A module's hidden atoms can be viewed as a special case of its output atoms which occur in no other modules. For simplicity, we consider only DLP-modules with no hidden atoms ($\mathcal{H} = \emptyset$), which we denote by a triple $(\Pi, \mathcal{I}, \O)$.

*Definition 2* (*Module Answer Set, (Janhunen et al. 2009)*)
We say that a set $X$ of atoms is a *(module) answer set* of a DLP-module $(\Pi, \mathcal{I}, \O)$ if $X$ is an answer set of $\Pi \cup \{p \mid p \in (\mathcal{I} \cap X)\}$.

The role of input atoms can be simulated using choice rules. A choice rule $\{p\} \leftarrow \imath Body$ is understood as shorthand for $p; \imath not\ p \leftarrow \imath Body$ (Lee et al. 2008). The following lemma shows how module answer sets can be alternatively characterized in terms of choice rules.

*Lemma 1*
$X$ is a module answer set of $(\Pi, \mathcal{I}, \O)$ iff $X$ is an answer set of $\Pi \cup \{\{p\} \leftarrow \mid p \in \mathcal{I}\}$.

*Definition 3* (*Dependency Graph of a DLP-Module*)
The *dependency graph* of a DLP-module $\mathbf{\Pi} = (\Pi, \mathcal{I}, \O)$, denoted by DG$[\Pi; \O]$, is the directed graph that

- has all members of $\O$ as its vertices, and
- has edges from each $a_i$ $(1 \leq i \leq n)$ to each $b_j$ $(1 \leq j \leq m)$ for each rule (3) in $\Pi$.

It is clear that this definition is a special case of Definition 1.

*Definition 4* (*Joinability of DLP-modules*)
Two DLP-modules $\mathbf{\Pi_1} = (\Pi_1, \mathcal{I}_1, \O_1)$ and $\mathbf{\Pi_2} = (\Pi_2, \mathcal{I}_2, \O_2)$ are called *joinable* if

- $\O_1 \cap \O_2 = \emptyset$,
- each strongly connected component of DG$[\Pi_1 \cup \Pi_2;\ \O_1 \O_2]$ is either a subset of $\O_1$ or a subset of $\O_2$,
- each rule in $\Pi_1$ ($\Pi_2$, respectively) whose head is not disjoint with $\O_2$ ($\O_1$, respectively) occurs in $\Pi_2$ ($\Pi_1$, respectively).

*Definition 5* (*Join of DLP-modules*)
For any modules $\mathbf{\Pi_1} = (\Pi_1, \mathcal{I}_1, \mathcal{O}_1)$ and $\mathbf{\Pi_2} = (\Pi_2, \mathcal{I}_2, \mathcal{O}_2)$ that are joinable, the *join* of $\mathbf{\Pi}_1$ and $\mathbf{\Pi}_2$, denoted by $\mathbf{\Pi}_1 \sqcup \mathbf{\Pi}_2$, is defined to be the DLP-module

$$(\Pi_1 \cup \Pi_2,\ (\mathcal{I}_1 \cup \mathcal{I}_2) \setminus (\mathcal{O}_1 \cup \mathcal{O}_2),\ \mathcal{O}_1 \cup \mathcal{O}_2)\ .$$

Informally, the join of two DLP-modules corresponds to the union of their programs, and defines all atoms that are defined by either module.

Given sets of atoms $X_1$, $X_2$, and $A$, we say that $X_1$ and $X_2$ are *A-compatible* if $X_1 \cap A = X_2 \cap A$. As demonstrated by Janhunen et al. (2009), given a program composed of a series of joinable DLP-modules, it is possible to consider each DLP-module contained in a program separately, evaluate them, and compose the resulting compatible answer sets in order to obtain the answer sets of the complete program. This notion is presented in Theorem 2, which is a reformulation of the main theorem (Theorem 5.7) from (Janhunen et al. 2009).

*Theorem 2* (*Module Theorem for DLPs*)
Let $\mathbf{\Pi}_1 = (\Pi_1, \mathcal{I}_1, \mathcal{O}_1)$ and $\mathbf{\Pi}_2 = (\Pi_2, \mathcal{I}_2, \mathcal{O}_2)$ be DLP-modules that are joinable, and let $X_1$ and $X_2$ be $((\mathcal{I}_1 \cup \mathcal{O}_1) \cap (\mathcal{I}_2 \cup \mathcal{O}_2))$-compatible sets of atoms. The set $X_1 \cup X_2$ is a module answer set of $\mathbf{\Pi}_1 \sqcup \mathbf{\Pi}_2$ iff $X_1$ is a module answer set of $\mathbf{\Pi}_1$ and $X_2$ is a module answer set of $\mathbf{\Pi}_2$.

## 3 A Generalization of the Splitting Theorem by Ferraris *et al.*

The module theorem (Theorem 2) and the splitting theorem (Theorem 1) resemble each other. When we restrict attention to propositional logic program $F$, the intensional predicates $\mathbf{p}$ in $\mathrm{SM}[F;\mathbf{p}]$ correspond to output atoms in the corresponding module. Though not explicit in the notation $\mathrm{SM}[F;\mathbf{p}]$, the predicates that are not in $\mathbf{p}$ behave like input atoms in the corresponding module. Also, the joinability condition in Definition 4 appears similar to the splitting condition in Theorem 1, but with one exception: the last clause in the definition of joinability (Definition 4) does not have a counterpart in the splitting theorem. The module theorem allows us to join two DLP-modules $\mathbf{\Pi}_1 = (\Pi_1, \mathcal{I}_1, \mathcal{O}_1)$ and $\mathbf{\Pi}_2 = (\Pi_2, \mathcal{I}_2, \mathcal{O}_2)$ even when $\Pi_1$ has a rule whose head contains an output atom in $\mathcal{O}_2$ as long as that rule is also in $\Pi_2$.

Indeed, this difference yields the splitting theorem less generally applicable than the module theorem in some cases. For example, the module theorem (Theorem 2) allows us to join

$$\begin{aligned} &\mathbf{\Pi}_1 = (\{p \vee q \leftarrow r. \quad s \leftarrow .\},\ \{q, r\},\ \{p, s\}) \text{ and} \\ &\mathbf{\Pi}_2 = (\{p \vee q \leftarrow r. \quad t \leftarrow .\},\ \{p, r\},\ \{q, t\}) \end{aligned} \tag{4}$$

into

$$\mathbf{\Pi} = (\{p \vee q \leftarrow r. \quad s \leftarrow . \quad t \leftarrow .\},\ \{r\},\ \{p, q, s, t\}).$$

On the other hand, the splitting theorem (Theorem 1), as presented in Section 2.2,

is not as general in this regard. It does not allow us to justify that

$$\mathrm{SM}[(r \rightarrow p \vee q) \wedge s;\ p, s] \wedge \mathrm{SM}[(r \rightarrow p \vee q) \wedge t;\ q, t] \tag{5}$$

is equivalent to

$$\mathrm{SM}[(r \rightarrow p \vee q) \wedge s \wedge t;\ p, q, s, t]\ . \tag{6}$$

because, for instance, $r \rightarrow p \vee q$ in the first conjunctive term of (5) is not negative on $\{q, t\}$.

In order to close the gap, we next extend the splitting theorem to allow a *partial split*, which allows an overlapping sentence, such as $r \rightarrow p \vee q$ in the above example, in both component formulas.

*Theorem 3* (*Extension of the Splitting Theorem*)
Let $F$, $G$, $H$ be first-order sentences, and let $\mathbf{p}$, $\mathbf{q}$ be finite lists of distinct predicate constants. If

(a) each strongly connected component of $\mathrm{DG}[F \wedge G \wedge H;\ \mathbf{pq}]$ is a subset of $\mathbf{p}$ or a subset of $\mathbf{q}$,
(b) $F$ is negative on $\mathbf{q}$, and
(c) $G$ is negative on $\mathbf{p}$

then

$$\mathrm{SM}[F \wedge G \wedge H;\ \mathbf{pq}] \leftrightarrow \mathrm{SM}[F \wedge H;\ \mathbf{p}] \wedge \mathrm{SM}[G \wedge H;\ \mathbf{q}]$$

is logically valid.

It is clear that Theorem 1 is a special case of Theorem 3 (take $H$ to be $\top$). Unlike in (Ferraris et al. 2009) we do not require $\mathbf{p}$ and $\mathbf{q}$ to be disjoint from each other.

Getting back to the example above, according to the extended splitting theorem, (6) is equivalent to (5) (Take $H$ to be $r \rightarrow p \vee q$).

## 4 Module Theorem for General Theory of Stable Models

### *4.1 Statement of the Theorem*

In this section, we present a new formulation of the module theorem that is applicable to first-order formulas under the stable model semantics.

As a step towards this end, we first define the notion of a partial interpretation. Given a signature $\sigma$ and its subset $\mathbf{c}$, by a $\mathbf{c}$-*partial interpretation* of $\sigma$, we mean an interpretation of $\sigma$ restricted to $\mathbf{c}$. Clearly, a $\sigma$-partial interpretation of $\sigma$ is simply an interpretation of $\sigma$. By an *Herbrand* $\mathbf{c}$-partial interpretation of $\sigma$, we mean an Herbrand interpretation of $\sigma$ restricted to $\mathbf{c}$.

We say that a $\mathbf{c}_1$-partial interpretation $I_1$ and a $\mathbf{c}_2$-partial interpretation $I_2$ of the same signature $\sigma$ are *compatible* if their universes are the same, and $c^{I_1} = c^{I_2}$ for every common constant $c$ in $\mathbf{c}_1 \cap \mathbf{c}_2$. For such compatible partial interpretations $I_1$ and $I_2$, we define the *union* of $I_1$ and $I_2$, denoted by $I_1 \cup I_2$, to be the $(\mathbf{c_1} \cup \mathbf{c_2})$-partial interpretation of $\sigma$ such that (*i*) $|I_1 \cup I_2| = |I_1| = |I_2|$ [2], (*ii*) $c^{I_1 \cup I_2} = c^{I_1}$ for every constant $c$ in $\mathbf{c}_1$, and (*iii*) $c^{I_1 \cup I_2} = c^{I_2}$ for every constant $c$ in $\mathbf{c}_2$.

[2] $|I|$ denotes the universe of the interpretation $I$.

Next we introduce a first-order analog to DLP-modules, which we refer to as *first-order modules*, and define a method of composing multiple such constructs similar to the join operation for DLP-modules. By $\imath pr(F)$ we denote the set of all predicate constants occurring in $F$. A *(first-order) module* $\mathbf{F}$ of a signature $\sigma$ is a triple $(F, \mathcal{I}, \emptyset)$, where $F$ is a first-order sentence of $\sigma$, and $\mathcal{I}$ and $\emptyset$ are disjoint lists of distinct predicate constants of $\sigma$ such that $\imath pr(F) \subseteq (\mathcal{I} \cup \emptyset)$. Intuitively, $\mathcal{I}$ and $\emptyset$ denote, respectively, the sets of non-intensional (input) and intensional (output) predicates considered by $F$.

*Definition 6* (*Module Stable Model*)
We say that an interpretation $I$ is a *(module) stable model* of a module $\mathbf{F} = (F, \mathcal{I}, \emptyset)$ if $I \models \mathrm{SM}[F; \emptyset]$. We understand $\mathrm{SM}[\mathbf{F}]$ as shorthand for $\mathrm{SM}[F; \emptyset]$.

*Definition 7* (*Joinability of First-Order Modules*)
Two first-order modules $\mathbf{F}_1 = (F_1 \wedge H,\ \mathcal{I}_1,\ \emptyset_1)$ and $\mathbf{F}_2 = (F_2 \wedge H,\ \mathcal{I}_2,\ \emptyset_2)$ are called *joinable* if

- $\emptyset_1 \cap \emptyset_2 = \emptyset$,
- each strongly connected component of $\mathrm{DG}[F_1 \wedge F_2 \wedge H;\ \emptyset_1 \cup \emptyset_2]$ is either a subset of $\emptyset_1$ or a subset of $\emptyset_2$,
- $F_1$ is negative on $\emptyset_2$, and
- $F_2$ is negative on $\emptyset_1$.

*Definition 8* (*Join of First-Order modules*)
For any modules $\mathbf{F}_1 = (F_1 \wedge H,\ \mathcal{I}_1,\ \emptyset_1)$ and $\mathbf{F}_2 = (F_2 \wedge H,\ \mathcal{I}_2,\ \emptyset_2)$ that are joinable, the *join* of $\mathbf{F}_1$ and $\mathbf{F}_2$, denoted by $\mathbf{F}_1 \sqcup \mathbf{F}_2$, is defined to be the first-order module

$$(F_1 \wedge F_2 \wedge H,\ (\mathcal{I}_1 \cup \mathcal{I}_2) \setminus (\emptyset_1 \cup \emptyset_2),\ \emptyset_1 \cup \emptyset_2)\ .$$

It is not difficult to check that this definition is a proper generalization of Definition 4.

As with DLP-modules, the join operation for first-order modules is both commutative and associative.

*Proposition 1* (*Commutativity and Associativity of Join*)
For any first-order modules $\mathbf{F}_1$, $\mathbf{F}_2$, and $\mathbf{F}_3$, the following properties hold:

- $\mathbf{F}_1 \sqcup \mathbf{F}_2$ is defined iff $\mathbf{F}_2 \sqcup \mathbf{F}_1$ is defined.
- $\mathrm{SM}[\mathbf{F}_1 \sqcup \mathbf{F}_2]$ is equivalent to $\mathrm{SM}[\mathbf{F}_2 \sqcup \mathbf{F}_1]$.
- $(\mathbf{F}_1 \sqcup \mathbf{F}_2) \sqcup \mathbf{F}_3$ is defined iff $\mathbf{F}_1 \sqcup (\mathbf{F}_2 \sqcup \mathbf{F}_3)$ is defined.
- $\mathrm{SM}[(\mathbf{F}_1 \sqcup \mathbf{F}_2) \sqcup \mathbf{F}_3]$ is equivalent to $\mathrm{SM}[\mathbf{F}_1 \sqcup (\mathbf{F}_2 \sqcup \mathbf{F}_3)]$.

The following theorem is an extension of Theorem 2 to the general theory of stable models. Given a first-order formula $F$, by $\mathbf{c}(F)$ we denote the set of all object, function and predicate constants occurring in $F$.

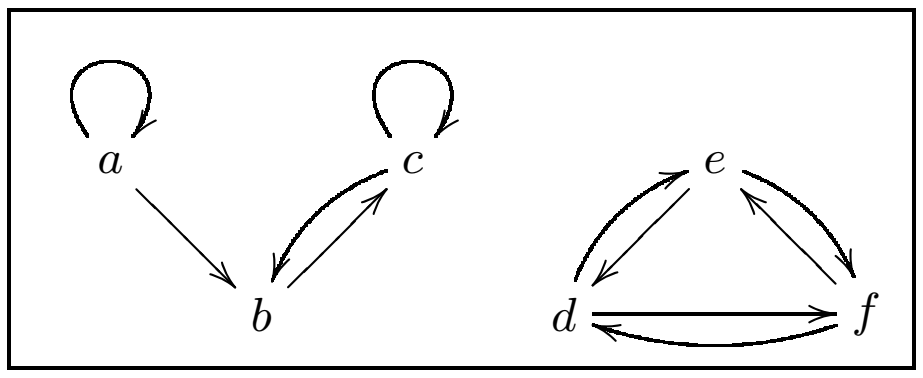

Fig. 1. A Simple Graph

*Theorem 4* (*Module Theorem for General Theory of Stable Models*)

Let $\mathbf{F}_1 = (F_1, \mathcal{I}_1, \mathcal{O}_1)$ and $\mathbf{F}_2 = (F_2, \mathcal{I}_2, \mathcal{O}_2)$ be first-order modules of a signature $\sigma$ that are joinable, and, for $i = 1, 2$, let $\mathbf{c}_i$ be a subset of $\sigma$ that contains $\mathbf{c}(F_i) \cup \mathcal{O}_i$, and let $I_i$ be a $\mathbf{c}_i$-partial interpretation of $\sigma$. If $I_1$ and $I_2$ are compatible with each other, then

$$I_1 \cup I_2 \models \text{SM}[\mathbf{F}_1 \sqcup \mathbf{F}_2] \qquad \text{iff} \qquad I_1 \models \text{SM}[\mathbf{F}_1] \text{ and } I_2 \models \text{SM}[\mathbf{F}_2] \ .$$

It is clear that when $\sigma = \mathbf{c}_1 = \mathbf{c}_2$, Theorem 4 reduces to Theorem 3.

Also, it is not difficult to check that Theorem 4 reduces to Theorem 2 when $\mathbf{F}_1$ and $\mathbf{F}_2$ represent DLP-modules, $\mathbf{c}_1$ is $\mathcal{I}_1 \cup \mathcal{O}_1$, and $\mathbf{c}_2$ is $\mathcal{I}_2 \cup \mathcal{O}_2$.

## 5 Example: Analyzing RASPL-1 Programs Using Module Theorem

As an example of Theorem 4, consider the problem of locating non-singleton cliques within a graph, such as the one shown in Figure 1, that are reachable from a pre-specified node. This problem can be divided into three essential parts: (*i*) fixing the graph, (*ii*) determining the reachable subgraph, and (*iii*) locating cliques within that subgraph.

We can describe the graph shown in Figure 1 in the language of RASPL-1 (Lee et al. 2008), which is essentially a fragment of the general theory of stable models in logic programming syntax. We assume that $\sigma$ is an underling signature. The program below lists the vertices and the edges using predicates `vertex` and `edge`, and assigns the starting vertex using `at` predicate.

$$\begin{array}{llllll}
\texttt{vertex}(a). & \texttt{vertex}(b). & \texttt{vertex}(c). & \texttt{vertex}(d). & \texttt{vertex}(e). & \texttt{vertex}(f). \\
\texttt{edge}(a,a). & \texttt{edge}(a,b). & \texttt{edge}(b,c). & \texttt{edge}(c,b). & \texttt{edge}(c,c). & \texttt{edge}(d,e). \\
\texttt{edge}(d,f). & \texttt{edge}(e,d). & \texttt{edge}(e,f). & \texttt{edge}(f,d). & \texttt{edge}(f,e). & \texttt{at}(a).
\end{array} \tag{7}$$

The first-order module $\mathbf{F}_G$ is $(F_G, \emptyset, \{\texttt{vertex}, \texttt{edge}, \texttt{at}\})$, where $F_G$ is the FOL-representation of program (7), which is the conjunction of all the atoms. Let $I_G$ be the following Herbrand $\mathbf{c}(F_G)$-partial interpretation of $\sigma$ that satisfies SM$[\mathbf{F}_G]$.

$$\begin{aligned}
\texttt{vertex}^{I_G} &= \{\texttt{a},\texttt{b},\texttt{c},\texttt{d},\texttt{e},\texttt{f}\},\\
\texttt{edge}^{I_G} &= \{(\texttt{a},\texttt{a}),(\texttt{a},\texttt{b}),(\texttt{b},\texttt{c}),(\texttt{c},\texttt{b}),\\
&\qquad (\texttt{c},\texttt{c}),(\texttt{d},\texttt{e}),(\texttt{d},\texttt{f}),(\texttt{e},\texttt{d}),(\texttt{e},\texttt{f}),(\texttt{f},\texttt{d}),(\texttt{f},\texttt{e})\}, \text{and}\\
\texttt{at}^{I_G} &= \{\texttt{a}\}.
\end{aligned}$$

The following program describes the reachable vertices by the predicate `reachable`, which is defined using `edge` and `at`.

$$\begin{aligned}
&\texttt{reachable}(X) \leftarrow \texttt{at}(X).\\
&\texttt{reachable}(Y) \leftarrow \texttt{reachable}(X), \texttt{edge}(X,Y).
\end{aligned} \tag{8}$$

The first-order module $\mathbf{F}_R$ is $(F_R, \{\texttt{edge},\texttt{at}\}, \{\texttt{reachable}\})$, where $F_R$ is the FOL-representation of program (8). Let $I_R$ be the following Herbrand $\mathbf{c}(F_R)$-partial interpretation of $\sigma$ that satisfies SM$[\mathbf{F}_G]$, which is compatible with $I_G$.

$$\begin{aligned}
\texttt{edge}^{I_R} &= \{(\texttt{a},\texttt{a}),(\texttt{a},\texttt{b}),(\texttt{b},\texttt{c}),(\texttt{c},\texttt{b}),\\
&\qquad (\texttt{c},\texttt{c}),(\texttt{d},\texttt{e}),(\texttt{d},\texttt{f}),(\texttt{e},\texttt{d}),(\texttt{e},\texttt{f}),(\texttt{f},\texttt{d}),(\texttt{f},\texttt{e})\},\\
\texttt{at}^{I_R} &= \{\texttt{a}\}, \text{ and}\\
\texttt{reachable}^{I_R} &= \{\texttt{a},\texttt{b},\texttt{c}\}.
\end{aligned}$$

Finally, the following program describes non-singleton cliques reachable from vertex $a$ by `in_clique`, which is defined using `edge` and `reachable`:

$$\begin{aligned}
&\{\texttt{in_clique(X)}\} \leftarrow \texttt{reachable}(X)\\
&\leftarrow \texttt{in_clique}(X),\ \texttt{in_clique}(Y),\ not\ \texttt{edge}(X,Y),\ X \neq Y\\
&\leftarrow not\ 2\{X : \texttt{in_clique}(X)\}.
\end{aligned} \tag{9}$$

In RASPL-1, expression $b\{\mathbf{x} : F(\mathbf{x})\}$, where $b$ is a positive integer, $\mathbf{x}$ is a list of object variables, and $F(\mathbf{x})$ is a conjunction of literals, stands for the first-order formula

$$\exists \mathbf{x}^1 \dots \mathbf{x}^b \left[ \bigwedge_{1\le i\le b} F(\mathbf{x}^i) \wedge \bigwedge_{1\le i<j\le b} \neg(\mathbf{x}^i = \mathbf{x}^j) \right],$$

where $\mathbf{x}^1,\dots,\mathbf{x}^b$ are lists of new object variables of the same length as $\mathbf{x}$. For any lists of variables $\mathbf{x} = (x_1,\dots,x_n)$ and $\mathbf{y} = (y_1,\dots,y_n)$ of the same length, $\mathbf{x} = \mathbf{y}$ stands for $x_1 = y_1 \wedge \cdots \wedge x_n = y_n$.

The first-order module $\mathbf{F}_C$ is $(F_C, \{\texttt{reachable},\texttt{edge}\}, \{\texttt{in_clique}\})$, where $F_C$ is the following FOL-representation of RASPL-1 program (9):

$$\begin{aligned}
&\forall X(\texttt{reachable}(X) \rightarrow (\texttt{in_clique}(X) \vee \neg\texttt{in_clique}(X)))\\
&\quad \wedge \forall XY(\texttt{in_clique}(X) \wedge \texttt{in_clique}(Y) \wedge \neg\texttt{edge}(X,Y) \wedge X \neq Y \ \rightarrow\ \bot)\\
&\quad \wedge (\neg\exists XY(\texttt{in_clique}(X) \wedge \texttt{in_clique}(Y) \wedge X \neq Y) \rightarrow \bot)\ .
\end{aligned}$$

Let $I_C$ be the following Herbrand $\mathbf{c}(F_C)$-partial interpretation of $\sigma$ that satisfies

SM[$\mathbf{F}_C$], which is compatible with $I_G$ and $I_R$.

$$\begin{aligned}\texttt{edge}^{I_C} &= \{(\texttt{a},\texttt{a}),(\texttt{a},\texttt{b}),(\texttt{b},\texttt{c}),(\texttt{c},\texttt{b}),\\ &\qquad (\texttt{c},\texttt{c}),(\texttt{d},\texttt{e}),(\texttt{d},\texttt{f}),(\texttt{e},\texttt{d}),(\texttt{e},\texttt{f}),(\texttt{f},\texttt{d}),(\texttt{f},\texttt{e})\},\\ \texttt{reachable}^{I_C} &= \{\texttt{a},\texttt{b},\texttt{c}\}, \text{ and}\\ \texttt{in_clique}^{I_C} &= \{\texttt{b},\texttt{c}\}.\end{aligned}$$

Clearly, $\mathbf{F}_G$, $\mathbf{F}_R$, and $\mathbf{F}_C$ are joinable. In accordance with Theorem 4, the union of the partial interpretations $I_G \cup I_R \cup I_C$ is a partial interpretation of $\sigma$ that satisfies SM[$\mathbf{F}_G \sqcup \mathbf{F}_R \sqcup \mathbf{F}_C$].

## 6 Modules That Can Be Incrementally Assembled

### *6.1 Review: Incremental Modularity by Gebser et al.*

In this section, we present a reformulation of the theory behind the system ICLINGO, which was developed to allow for incremental grounding and solving of answer set programs. We follow the enhancement given in (Gebser et al. 2011) with a slight deviation. Most notably, we do not restrict attention to nondisjunctive logic programs, but limit attention to offline programs for simplicity.

Given a disjunctive program $\Pi$ of a signature $\sigma$, by $\text{Ground}_\sigma(\Pi)$ we denote the ground program obtained from $\Pi$ by replacing object variables with ground terms in the Herbrand Universe of $\sigma$. If $\Pi$ is ground, then the *projection* of $\Pi$ onto a set $X$ of ground atoms, denoted by $\Pi|_X$, is defined to be the program obtained from $\Pi$ by removing all rules (3) in $\Pi$ that contain some $b_i$ not in $X$, and then removing all occurrences of $\mathit{not}\ c_j$ such that $c_j$ is not in $X$ from the remaining rules. By $\mathit{head}(\Pi)$ we denote the set of all atoms that occur in the head of a rule in $\Pi$.

*Definition 9* (*DLP-Module Instantiation*)
Given a disjunctive program $\Pi$, and a set of ground atoms $\mathcal{I}$, Gebser et al. (2011) define the *DLP-module instantiation* of $\Pi$ w.r.t. $\mathcal{I}$, denoted by $\mathit{DM}(\Pi,\mathcal{I})$, to be the DLP-module $(\text{Ground}_\sigma(\Pi)|_{\mathcal{I}\cup\mathcal{O}}, \mathcal{I}, \mathcal{O})$, where $\mathcal{O}$ is $\mathit{head}\big(\text{Ground}_\sigma(\Pi)|_{\mathcal{I}\cup \mathit{head}(\text{Ground}_\sigma(\Pi))}\big)$.

For example, Figure 2 shows a simple program and its DLP-module instantiation w.r.t. $\{l, t\}$.

An *incrementally parameterized program* $\Pi[t]$ is a program which may contain atoms of the form $a_{f(t)}(\mathbf{x})$, called *incrementally parameterized atoms*, where $t$ is an *incremental step counter*, and $f(t)$ is some arithmetic function involving $t$. Given such a program $\Pi[t]$, its *incremental instantiation* at some nonnegative integer $i$, which we denote by $\Pi[i]$, is defined to be the program obtained by replacing all occurrences of atoms $a_{f(t)}(\mathbf{x})$ with an atom $a_v(\mathbf{x})$, where $v$ is the result of evaluating

$$\left.\begin{array}{l} n \leftarrow t\\ p \leftarrow q, t\\ q \leftarrow r, \mathit{not}\ s\\ r \leftarrow m \end{array}\right\} \underset{I=\{l,t\}}{\longmapsto} \left(\begin{array}{l} n \leftarrow t\\ p \leftarrow q, t\end{array}, \{l,t\}, \{n,p,q\}\right)$$

Fig. 2. DLP-module instantiation of a simple program

$f(i)$. For example, let $\Pi = \{p_{t+1}(x) \leftarrow p_t(x), \imath not\ q(x)\}$. The program $\Pi[2]$ is then $\{p_3(x) \leftarrow p_2(x), \imath not\ q(x)\}$.

Gebser et al. (2011) define an *incremental logic program* to be a triple $\langle B, P[t], Q[t]\rangle$, where $B$ is a disjunctive logic program, and $P[t]$, $Q[t]$ are incrementally parameterized disjunctive logic programs. Informally, $B$ is the *base* program component, which describes static knowledge; $P[t]$ is the *cumulative* program component, which contains information regarding every step $t$ that should be accumulated during execution; $Q[t]$ is the *volatile query* program component, containing constraints or information regarding the final step.

We assume a partial order $\prec$ on

$$\begin{aligned}\{\text{Ground}_\sigma(B), &\text{Ground}_\sigma(P[1]), \text{Ground}_\sigma(P[2]), \dots, \\ &\text{Ground}_\sigma(Q[1]), \text{Ground}_\sigma(Q[2]), \dots\}\end{aligned} \tag{10}$$

such that

- $\text{Ground}_\sigma(B) \prec \text{Ground}_\sigma(P[1]) \prec \text{Ground}_\sigma(P[2]) \prec \dots$;
- $\text{Ground}_\sigma(P[i]) \prec \text{Ground}_\sigma(Q[i])$ for $i \geq 1$.

Given a DLP-module $\mathbf{P} = (\Pi, \mathcal{I}, \emptyset)$, by $\imath Out(\mathbf{P})$ we denote $\emptyset$.

*Definition 10* (*Modular and Acyclic Logic Programs*)
An incremental logic program $\langle B, P[t], Q[t]\rangle$ is *modular* if the following DLP-modules are defined for every $k \geq 0$:

$$\begin{aligned}\mathbf{P}_0 &= \imath DM(B, \emptyset), \\ \mathbf{P}_i &= \mathbf{P}_{i-1} \sqcup \imath DM(P[i], \imath Out(\mathbf{P}_{i-1})), && (1 \leq i \leq k) \\ \mathbf{R}_k &= \mathbf{P}_k \sqcup \imath DM(Q[k], \imath Out(\mathbf{P}_k)),\end{aligned}$$

and is *acyclic* if, for each pair of programs $\Pi$, $\Pi'$ in (10) such that $\Pi \prec \Pi'$, we have that $\Pi$ contains no head atoms of $\Pi'$. [3]

Gebser et al. (2011) demonstrated that given a modular and acyclic incremental logic program $\langle B, P[t], Q[t]\rangle$ and some nonnegative integer $k$, we are able to evaluate each component DLP-module individually, and compose the results in order to obtain the answer sets of the complete module $\mathbf{R}_k$. They define the *$k$-expansion* $R_k$ of the incremental logic program as

$$B \cup P[1] \cup \dots \cup P[k] \cup Q[k] \ .$$

*Proposition 2*
(Gebser et al. 2011, Proposition 2) Let $\langle B, P[t], Q[t]\rangle$ be an incremental logic program of a signature $\sigma$ that is modular and acyclic, let $k$ be a nonnegative integer, and let $X$ be a subset of the output atoms of $\mathbf{R}_k$. Set $X$ is an answer set of the $k$-expansion $R_k$ of $\langle B, P[t], Q[t]\rangle$ if and only if $X$ is a (module) answer set of $\mathbf{R}_k$.

[3] The acyclicity condition corresponds to the special case of the "mutually revisable" condition in (Gebser et al. 2011) when there is no online component.

Proposition 2 tells us that the results of incrementally grounding and evaluating an incremental logic program are identical to the results of evaluating the entire $k$-expansion in the usual non-incremental fashion.

### *6.2 Incrementally Assembled First-Order Modules*

In this section, we consider an extension of the theory supporting system ICLINGO which allows for the consideration of first-order sentences by utilizing Theorem 4. This extension may be useful in analyzing non-ground RASPL-1 programs that describe dynamic domains.

Given a first-order sentence $F$, we define the *projection* of $F$ onto a set $\mathbf{p}$ of predicates, denoted by $F|_{\mathbf{p}}$, to be the first-order sentence obtained by replacing all occurrences of atoms of the form $q(t_1, \dots, t_n)$ in $F$ such that $q \in \imath pr(F) \setminus \mathbf{p}$ with $\bot$ and performing the following syntactic transformations recursively until no further transformations are possible:

$$
\begin{array}{llll}
\neg\bot \mapsto \top & \neg\top \mapsto \bot & & \\
\bot \wedge F \mapsto \bot & F \wedge \bot \mapsto \bot & \top \wedge F \mapsto F & F \wedge \top \mapsto F \\
\bot \vee F \mapsto F & F \vee \bot \mapsto F & \top \vee F \mapsto \top & F \vee \top \mapsto \top \\
\bot \rightarrow F \mapsto \top & F \rightarrow \top \mapsto \top & \top \rightarrow F \mapsto F & \\
\exists x \top \mapsto \top & \exists x \bot \mapsto \bot & \forall x \top \mapsto \top & \forall x \bot \mapsto \bot
\end{array}
$$

For example, consider the first-order sentence

$$\forall x(p(x) \rightarrow q(x)) \wedge (q(a) \wedge \neg p(a) \rightarrow r) \wedge \forall x(\neg q(x) \wedge t(x) \rightarrow s(x)) \ . \qquad (11)$$

The projection of (11) onto $\{q, r, s, t, m\}$ is

$$(q(a) \rightarrow r) \wedge \forall x(\neg q(x) \wedge t(x) \rightarrow s(x)) \ .$$

When we restrict attention to the case of propositional logic programs such that $\mathbf{p}$ contains at least the predicates occurring strictly positively in $F$, this notion coincides with the corresponding one in the previous section.

Similar to incrementally parameterized programs, we define an *incrementally parameterized* formula $F[t]$ to be a first-order formula which may contain incrementally parameterized atoms. For any nonnegative integer $i$, we define the *incremental instantiation* of $F$ at $i$, denoted by $F[i]$, to be the result of replacing all occurrences of incrementally parameterized atoms $a_{f(t)}(\mathbf{x})$ in $F[t]$ with an atom $a_v(\mathbf{x})$, where $v$ is the result of evaluating $f(i)$.

*Definition 11* (*First-Order Module Instantiation*)
For any first-order sentence $F$ and any set of (input) predicates $\mathcal{I}$, formula $F^0$ is defined as $F$, and $F^{i+1}$ is defined as $F^i|_{\mathcal{I} \cup head(F^i)}$, where $\imath head(F^i)$ denotes the set of all predicates occurring strictly positively in $F^i$. We define the *first-order module instantiation* of $F$ w.r.t. $\mathcal{I}$, denoted by $\imath FM(F, \mathcal{I})$, to be the first-order module

$$(F^\omega,\ \mathcal{I},\ \imath pr(F) \setminus \mathcal{I}),$$

where $F^\omega$ is the least fixpoint of the sequence $F^0, F^1, \dots$.

The idea of the simplification process is related to the fact that all predicates other than the ones in $\mathcal{I} \cup \imath head(F^i)$ have empty extents under the stable model semantics, which are equivalent to $\bot$ (Ferraris et al. 2011, Theorem 4). The process is guaranteed to lead to a fixpoint in a finite number of steps since $F$ is finite and $F^i|_{\mathcal{I} \cup head(F^i)}$ is shorter than $F^i$ in all cases except for the terminating case. It is not difficult to check that if $F$ is the FOL-representation of a ground disjunctive program $\Pi$, the first component $\text{Ground}_\sigma(\Pi)|_{\mathcal{I} \cup \mathcal{O}}$ in the definition of a DLP-module instantiation corresponds to $F^2$.

*Example 2*
Consider the propositional formula

$$F = (p \rightarrow q) \wedge (q \rightarrow r) \wedge (t \wedge \neg r \rightarrow s) \ .$$

and $\mathcal{I} = \{t, m\}$. The process of instantiation results in the following transformations on $F$:

$$\begin{array}{lll} & (p \rightarrow q) \wedge (q \rightarrow r) \wedge (t \wedge \neg r \rightarrow s). & F^0 (= F) \\ \Rightarrow & (q \rightarrow r) \wedge (t \wedge \neg r \rightarrow s). & F^1 \\ \Rightarrow & t \wedge \neg r \rightarrow s. & F^2 \\ \Rightarrow & t \rightarrow s. & F^3 \\ \Rightarrow & t \rightarrow s. & F^4 \end{array}$$

The resulting first-order module is then

$$\imath FM(F, \{t, m\}) = (t \rightarrow s, \quad \{t, m\}, \quad \{p, q, r, s\}).$$

This definition of an instantiation is different from the one by Gebser et al. (2011) even when we restrict attention to a finite propositional disjunctive program. First, we maximize the simplification done on the initial formula $F$ by repeatedly projecting it onto its head and input predicates, whereas Gebser et al. perform only the first two projections (i.e., $F^2$). Second, the list of output atoms are different. In our case all atoms occurring in $F$ that are not input atoms are assumed to be output atoms. The following example illustrates these differences.

*Example 3*
Recall the DLP-module instantiation in Figure 2. The first-order instantiation of (the FOL-representation of) the program w.r.t $\{l, t\}$ is $(t \rightarrow n,\ \{l, t\},\ \{m, n, p, q, r, s\})$.

While the two notions of instantiation are syntactically different, it can be shown that, given a propositional logic program $\Pi$ and sets of propositional atoms $I$ and $X$, $X$ is a module answer set of $\imath DM(\Pi, I)$ if and only if $X$ is a module answer set of $\imath FM(\Pi, I)$.

An *incremental first-order theory* is a triple $\langle B, P[t], Q[t] \rangle$ where $B$ is a first-order sentence, and $P[t]$ and $Q[t]$ are incrementally parameterized sentences.

The *k-expansion* of $\langle B, P[t], Q[t] \rangle$ is defined as

$$R_k = B \wedge P[1] \wedge \cdots \wedge P[k] \wedge Q[k].$$

It is clear that this coincides with the notion of $k$-expansion for incremental logic programs when we restrict attention to the common syntax.

We assume a partial order $\prec$ on

$$\{B, P[1], P[2], \ldots, Q[1], Q[2], \ldots\} \qquad (12)$$

as follows:

- $B \prec P[1] \prec P[2] \prec \ldots$ ;
- $P[i] \prec Q[i]$ for $i \geq 1$.

*Definition 12* (*Acyclic Incremental First-Order Theory*)
We say that an incremental first-order theory $\langle B, P[t], Q[t]\rangle$ is *acyclic* if, for every pair of formulas $F, G$ in (12) such that $F \prec G$, we have that $G$ is negative on $\imath pr(F)$.

This definition of acyclicity mirrors that of Gebser *et al.*'s (2011) in that it prevents predicates from occurring strictly positively in multiple sentences which are instantiated from the incremental theory. However, as shown in Proposition 3, it is unnecessary to check a condition similar to modularity for incremental first-order theories, as it is ensured by acyclicity.

Given a first-order module $\mathbf{F} = (F, \mathcal{I}, \emptyset)$, by $\imath Out(\mathbf{F})$ we denote $\emptyset$.

*Proposition 3* (*Modularity of Incremental Theory*)
If an incremental first-order theory $\langle B, P[t], Q[t]\rangle$ is acyclic, then the following modules are defined for all $k \geq 0$.

$$\begin{aligned} \mathbf{P}_0 &= \imath FM(B, \emptyset), \\ \mathbf{P}_i &= \mathbf{P}_{i-1} \sqcup \imath FM(P[i], \imath Out(\mathbf{P}_{i-1})), \qquad (1 \leq i \leq k) \\ \mathbf{R}_k &= \mathbf{P}_k \sqcup \imath FM(Q[k], \imath Out(\mathbf{P}_k)) \ . \end{aligned}$$

By applying Theorem 4, we can evaluate each component module independently and compose their results in order to obtain the stable models of $\mathbf{R}_k$.

*Proposition 4* (*Compositionality for Incremental First-Order Theories*)
Let $\langle B, P[t], Q[t]\rangle$ be an incremental first-order theory and let $\mathbf{R}_k$ be the module as defined in the statement of Proposition 3. For any nonnegative integer $k$,

$$\begin{aligned} I_B \cup I_{P[1]} \cup \cdots \cup I_{P[k]} \cup I_{Q[k]} &\models \mathrm{SM}[\mathbf{R}_k] \\ \text{iff} \quad & I_B \models \mathrm{SM}[\imath FM(B, \emptyset)] \\ & \text{and} \quad I_{P[1]} \models \mathrm{SM}[\imath FM(P[1], \imath Out(\mathbf{P}_0))] \\ & \text{and} \quad \ldots \qquad (13) \\ & \text{and} \quad I_{P[k]} \models \mathrm{SM}[\imath FM(P[k], \imath Out(\mathbf{P}_{k-1}))] \\ & \text{and} \quad I_{Q[k]} \models \mathrm{SM}[\imath FM(Q[k], \imath Out(\mathbf{P}_k))] \ . \end{aligned}$$

where $I_B$ ($I_{P[1]}, \ldots, I_{P[k]}, I_{Q[k]}$, respectively) is a $\mathbf{c}(B)$-partial interpretation ($\mathbf{c}(P[1])$, $\ldots, \mathbf{c}(P[k]), \mathbf{c}(Q[k])$-partial interpretation, respectively) such that $I_B, I_{P[1]}, \ldots, I_{P[k]}, I_{Q[k]}$ are pairwise compatible.

Given an acyclic incremental theory and a nonnegative integer $k$, the following proposition states that evaluating the individual modules and composing their results is equivalent to evaluating the $k$-expansion of the incremental theory.

*Proposition 5* (*Correctness of Incremental Assembly*)
Let $\langle B, P[t], Q[t]\rangle$ be an acyclic incremental theory, let $k$ be a nonnegative integer, let $R_k$ be the $k$-expansion of the incremental theory, and let $\mathbf{R}_k$ be the module as defined in Proposition 3. For any $\mathbf{c}$-partial interpretation $I$ such that $\mathbf{c} \supseteq \mathbf{c}(R_k)$, we have that

$$I \models \mathrm{SM}[R_k] \;\text{ iff }\; I \models \mathrm{SM}[\mathbf{R}_k].$$

## 7 Conclusion

Our extension of the module theorem to the general theory of stable models applies to non-ground logic programs containing choice rules, the count aggregate, and nested expressions. The extension is based on the new findings about the relationship between the module theorem and the splitting theorem. The proof of our module theorem[4] uses the splitting theorem as a building block so that a further generalization of the splitting theorem can be applied to generalize the module theorem as well. Indeed, the module theorem presented here can be extended to logic programs with arbitrary (recursive) aggregates, based on the extension of the splitting theorem to formulas with generalized quantifiers, recently presented in (Lee and Meng 2012). Based on the generalized module theorem, we reformulated and extended the theory of incremental answer set computation to the general theory of stable models, which can be useful in analyzing non-ground RASPL-1 programs that describe dynamic domains.

## Acknowledgements

We are grateful to Martin Gebser and Tomi Janhunen for useful discussions related to this paper. We are also grateful to the anonymous referees for their useful comments. This work was partially supported by the National Science Foundation under Grant IIS-0916116.

[4] Available in the online appendix.

## Appendix A Proofs

### *A.1 Splitting Lemma*

We use the splitting lemma (Ferraris et al. 2009) to prove a few theorems below.

*Splitting Lemma*
Let $F$ be a first-order sentence, and let $\mathbf{p}$, $\mathbf{q}$ be lists of distinct predicate constants. If each strongly connected component of $\mathrm{DG}[F;\mathbf{pq}]$ is a subset of $\mathbf{p}$ or a subset of $\mathbf{q}$ then

$$\mathrm{SM}[F;\mathbf{pq}] \quad \text{is equivalent to} \quad \mathrm{SM}[F;\mathbf{p}] \wedge \mathrm{SM}[F;\mathbf{q}]\ .$$

The statement is slightly more general than the one from (Ferraris et al. 2009) in that $\mathbf{p}$ and $\mathbf{q}$ are not required to be disjoint. The proof of this enhancement follows from the Version 3 of the Splitting Lemma from (Ferraris et al. 2009).

### *A.2 Proof of Lemma 1*

*Lemma 1*
$X$ is a module answer set of $(\Pi, \mathcal{I}, \emptyset)$ iff $X$ is an answer set of $\Pi \cup \{\{p\} \leftarrow \mid p \in \mathcal{I}\}$.

*Proof*

$$X \text{ is an answer set of } \Pi \cup \{p \leftarrow \mid p \in (\mathcal{I} \cap X)\}$$

iff

$$X \text{ is an answer set of } \Pi \cup \{p \leftarrow \mathit{not}\ \mathit{not}\ p \mid p \in \mathcal{I}\}$$

iff

$$X \text{ is an answer set of } \Pi \cup \{\{p\} \leftarrow \mid p \in \mathcal{I}\}\ .$$

The equivalence between the first and the second follows from the equivalence between the reducts of each program relative to $X$.

The equivalence between the second and third is because the transformation preserves strong equivalence. ∎

### A.3 Proof of Theorem 3

*Theorem 3*
Let $F$, $G$, $H$ be first-order sentences, and let $\mathbf{p}$, $\mathbf{q}$ be finite lists of distinct predicate constants. If

(a) each strongly connected component of $\mathrm{DG}[F \wedge G \wedge H;\ \mathbf{pq}]$ is a subset of $\mathbf{p}$ or a subset of $\mathbf{q}$,
(b) $F$ is negative on $\mathbf{q}$, and
(c) $G$ is negative on $\mathbf{p}$

then

$$\mathrm{SM}[F \wedge G \wedge H;\ \mathbf{pq}] \quad \text{is equivalent to} \quad \mathrm{SM}[F \wedge H;\ \mathbf{p}] \wedge \mathrm{SM}[G \wedge H;\ \mathbf{q}]\ .$$

*Proof*
By the Splitting Lemma above, $\mathrm{SM}[F \wedge G \wedge H;\ \mathbf{pq}]$ is equivalent to

$$\mathrm{SM}[F \wedge G \wedge H;\ \mathbf{p}] \wedge \mathrm{SM}[F \wedge G \wedge H;\ \mathbf{q}]\ .$$

Since $G$ is negative on $\mathbf{p}$, the first conjunctive term can be rewritten as

$$\mathrm{SM}[F \wedge H;\ \mathbf{p}] \wedge G\ . \tag{A1}$$

Similarly, the second conjunctive term can be rewritten as

$$\mathrm{SM}[G \wedge H;\ \mathbf{q}] \wedge F\ . \tag{A2}$$

It remains to observe that the second conjunctive term of each of the formulas (A1) and (A2) is entailed by the first conjunctive term of the other.
∎

### A.4 Proof of Proposition 1

*Proposition 1*
For any first-order modules $\mathbf{F}_1$, $\mathbf{F}_2$, and $\mathbf{F}_3$, the following properties hold:

- $\mathbf{F}_1 \sqcup \mathbf{F}_2$ is defined iff $\mathbf{F}_2 \sqcup \mathbf{F}_1$ is defined.
- $\mathrm{SM}[\mathbf{F}_1 \sqcup \mathbf{F}_2]$ is equivalent to $\mathrm{SM}[\mathbf{F}_2 \sqcup \mathbf{F}_1]$.
- $(\mathbf{F}_1 \sqcup \mathbf{F}_2) \sqcup \mathbf{F}_3$ is defined iff $\mathbf{F}_1 \sqcup (\mathbf{F}_2 \sqcup \mathbf{F}_3)$ is defined.
- $\mathrm{SM}[(\mathbf{F}_1 \sqcup \mathbf{F}_2) \sqcup \mathbf{F}_3]$ is equivalent to $\mathrm{SM}[\mathbf{F}_1 \sqcup (\mathbf{F}_2 \sqcup \mathbf{F}_3)]$.

*Proof*
Claims (a) and (b) follow immediately from the definitions.

We prove Claim (c). Let $\mathbf{F}_i = (F_i, \mathcal{I}_i, \mathbf{O}_i)$ for each $i \in \{1, 2, 3\}$ and without loss of generality assume that each $F_i$ is a conjunction of the form $F_{i,1} \wedge \cdots \wedge F_{i,k_i}$.

*From left to right:* Assume that $(\mathbf{F}_1 \sqcup \mathbf{F}_2) \sqcup \mathbf{F}_3$ is defined. Since $\mathbf{F}_1$ and $\mathbf{F}_2$ are joinable,

(i) $\mathbf{O}_1 \cap \mathbf{O}_2 = \emptyset$;
(ii) each conjunctive term of $F_1$ is negative on $\mathbf{O}_2$, or is one of the conjunctive terms of $F_2$;
(iii) each conjunctive term of $F_2$ is negative on $\mathbf{O}_1$, or is one of the conjunctive terms of $F_1$;
(iv) each strongly connected component of $\mathrm{DG}[F_1 \wedge F_2; \mathbf{O}_1\mathbf{O}_2]$ is a subset of $\mathbf{O}_1$ or a subset of $\mathbf{O}_2$.

Also, since $(\mathbf{F}_1 \sqcup \mathbf{F}_2)$ and $\mathbf{F}_3$ are joinable,

(v) $(\mathbf{O}_1 \cup \mathbf{O}_2) \cap \mathbf{O}_3 = \emptyset$;
(vi) each conjunctive term of $F_1 \wedge F_2$ is negative on $\mathbf{O}_3$, or is one of the conjunctive terms of $F_3$;
(vii) each conjunctive term of $F_3$ is negative on $\mathbf{O}_1 \cup \mathbf{O}_2$, or is one of the conjunctive terms of $F_1 \wedge F_2$;
(viii) each strongly connected component of $\mathrm{DG}[F_1 \wedge F_2 \wedge F_3; \mathbf{O}_1\mathbf{O}_2\mathbf{O}_3]$ is a subset of $\mathbf{O}_1 \cup \mathbf{O}_2$ or a subset of $\mathbf{O}_3$.

We first prove that $\mathbf{F}_2 \sqcup \mathbf{F}_3$ is defined.

(ix) From (v), it follows that $\mathbf{O}_2 \cap \mathbf{O}_3 = \emptyset$.
(x) From (vi), it follows that each conjunctive term of $F_2$ is negative on $\mathbf{O}_3$ or is one of the conjunctive terms of $F_3$.
(xi) We prove that each conjunctive term of $F_3$ is negative on $\mathbf{O}_2$ or is one of the conjunctive terms of $F_2$.
Consider any conjunctive term $C$ of $F_3$. By (vii), $C$ is negative on $\mathbf{O}_1 \cup \mathbf{O}_2$, or is one of the conjunctive terms of $F_1 \wedge F_2$.

— Case 1: $C$ is negative on $\mathbf{O}_1 \cup \mathbf{O}_2$. Clearly, it is negative on $\mathbf{O}_2$ as well.

— Case 2: $C$ is one of the conjunctive terms of $F_1 \wedge F_2$. If $C$ is one of the conjunctive terms of $F_2$, the claim trivially follows. If $C$ is one of the conjunctive terms of $F_1$, by (ii), it is either negative on $\emptyset_2$ or is one of the conjunctive terms of $F_2$. In either case, the claim follows.

(xii) We first prove that each strongly connected component of $\mathrm{DG}[F_1 \wedge F_2 \wedge F_3; \emptyset_1\emptyset_2\emptyset_3]$ is contained in only one of $\emptyset_1$, $\emptyset_2$ or $\emptyset_3$, from which the fact that each strongly connected component of $\mathrm{DG}[F_2 \wedge F_3; \emptyset_2\emptyset_3]$ is contained in $\emptyset_2$ or $\emptyset_3$ follows, as $\mathrm{DG}[F_2 \wedge F_3; \emptyset_2\emptyset_3]$ is a subgraph of $\mathrm{DG}[F_1 \wedge F_2 \wedge F_3; \emptyset_1\emptyset_2\emptyset_3]$.
By (i) and (v), $\emptyset_1$, $\emptyset_2$ and $\emptyset_3$ are pairwise disjoint. Consider any strongly connected component $S$ of $\mathrm{DG}[F_1 \wedge F_2 \wedge F_3; \emptyset_1\emptyset_2\emptyset_3]$. By (viii) $S$ is a subset of $\emptyset_1 \cup O_2$ or a subset of $\emptyset_3$. Assume that $S$ is a subset of $\emptyset_1 \cup \emptyset_2$. Clearly, $S$ is also a strongly connected component of $\mathrm{DG}[F_1 \wedge F_2 \wedge F_3; \emptyset_1\emptyset_2]$. In view of (vii), $\mathrm{DG}[F_1 \wedge F_2 \wedge F_3; \emptyset_1\emptyset_2]$ is the same as $\mathrm{DG}[F_1 \wedge F_2; \emptyset_1\emptyset_2]$, so that $S$ is a strongly connected component of $\mathrm{DG}[F_1 \wedge F_2; \emptyset_1\emptyset_2]$ as well. By (iv) $S$ is contained in $\emptyset_1$ or $\emptyset_2$.

We now prove that $\mathbf{F_1} \sqcup (\mathbf{F_2} \sqcup \mathbf{F_3})$ is defined.

- From (i) and (v), it follows that $\emptyset_1 \cap (\emptyset_2 \cup O_3) = \emptyset$;
- From (ii) and (vi), it follows that each conjunctive term of $F_1$ is negative on $\emptyset_2 \cup \emptyset_3$ or is one of the conjunctive terms of $F_2 \wedge F_3$;
- From (iii) and (vii), it follows that each conjunctive term of $F_2 \wedge F_3$ is negative on $\emptyset_1$ or is one of the conjunctive terms of $F_1$;
- From the claim proven in (viii), it follows that each strongly connected component of $\mathrm{DG}[F_1 \wedge F_2 \wedge F_3; \emptyset_1\emptyset_2\emptyset_3]$ is contained in $\emptyset_1$ or $\emptyset_2 \cup \emptyset_3$.

*From right to left:* Assume that $\mathbf{F}_1 \sqcup (\mathbf{F}_2 \sqcup \mathbf{F}_3)$ is defined. By Claim (a), $(\mathbf{F}_2 \sqcup \mathbf{F}_3) \sqcup \mathbf{F}_1$ is defined, and then $(\mathbf{F}_3 \sqcup \mathbf{F}_2) \sqcup \mathbf{F}_1$ is defined. By the first part of Claim (c) that was proven, $\mathbf{F}_3 \sqcup (\mathbf{F}_2 \sqcup \mathbf{F}_1)$ is defined, and then by applying Claim (a) twice, we have that $(\mathbf{F}_1 \sqcup \mathbf{F}_2) \sqcup \mathbf{F}_3$ is defined.

We now prove Claim (d). Using Theorem 4 and Claim (c),

$$\begin{aligned}\mathrm{SM}[(\mathbf{F}_1 \sqcup \mathbf{F}_2) \sqcup \mathbf{F}_3] &\Leftrightarrow \mathrm{SM}[\mathbf{F}_1 \sqcup \mathbf{F}_2] \wedge \mathrm{SM}[\mathbf{F}_3] \\ &\Leftrightarrow \mathrm{SM}[\mathbf{F}_1] \wedge \mathrm{SM}[\mathbf{F}_2] \wedge \mathrm{SM}[\mathbf{F}_3] \\ &\Leftrightarrow \mathrm{SM}[\mathbf{F}_1] \wedge \mathrm{SM}[\mathbf{F}_2 \sqcup \mathbf{F}_3] \\ &\Leftrightarrow \mathrm{SM}[\mathbf{F}_1 \sqcup (\mathbf{F}_2 \sqcup \mathbf{F}_3)] \ .\end{aligned}$$

∎

### *A.5 Proof of Theorem 4*

*Theorem 4*
Let $\mathbf{F}_1 = (F_1, \mathcal{I}_1, \emptyset_1)$ and $\mathbf{F}_2 = (F_2, \mathcal{I}_2, \emptyset_2)$ be first-order modules of a signature $\sigma$ that are joinable, and, for $i = 0, 1$, let $\mathbf{c}_i$ be a subset of $\sigma$ that contains $\mathbf{c}(F_i) \cup \emptyset_i$, and let $I_i$ be a $\mathbf{c}_i$-partial interpretation of $\sigma$. If $I_1$ and $I_2$ are compatible with each other, then

$$I_1 \cup I_2 \models \mathrm{SM}[\mathbf{F}_1 \sqcup \mathbf{F}_2] \quad \text{iff} \qquad I_1 \models \mathrm{SM}[\mathbf{F}_1] \ \text{ and } \ I_2 \models \mathrm{SM}[\mathbf{F}_2]\ .$$

*Proof*

Let us identify $\mathbf{F}_1$ with $(F_1' \wedge H, \mathcal{I}_1, \mathcal{O}_1)$ and $\mathbf{F}_2$ with $(F_2' \wedge H, \mathcal{I}_2, \mathcal{O}_2)$ as in the definition of join (Definition 8).

By definition $\mathrm{SM}[\mathbf{F}_1 \sqcup \mathbf{F}_2]$ is $\mathrm{SM}[F_1' \wedge F_2' \wedge H;\ \mathcal{O}_1 \cup \mathcal{O}_2]$. By Theorem 3,

$$\begin{array}{l} I_1 \cup I_2 \models \mathrm{SM}[F_1' \wedge F_2' \wedge H;\ \mathcal{O}_1 \cup \mathcal{O}_2] \text{ iff} \\ \qquad I_1 \cup I_2 \models \mathrm{SM}[F_1' \wedge H;\ \mathcal{O}_1] \text{ and } I_1 \cup I_2 \models \mathrm{SM}[F_2' \wedge H;\ \mathcal{O}_2] \end{array}$$

Clearly, $I_1 \cup I_2$ is compatible with $I_1$. Since $\mathbf{c}_1$ contains $\mathbf{c}(F_1' \wedge H) \cup \mathcal{O}_1$, it follows that $I_1 \cup I_2 \models \mathrm{SM}[F_1' \wedge H;\ \mathcal{O}_1]$ iff $I_1 \models \mathrm{SM}[F_1' \wedge H;\ \mathcal{O}_1]$. Similarly, $I_1 \cup I_2 \models \mathrm{SM}[F_2' \wedge H;\ \mathcal{O}_2]$ iff $I_2 \models \mathrm{SM}[F_2' \wedge H;\ \mathcal{O}_2]$. Consequently, the claim follows. ∎

### *A.6 Proof of Proposition 3*

*Lemma 2*

Let $\langle B, P[t], Q[t]\rangle$ be an incremental first-order theory, and let $\mathbf{P}_i$ and $\mathbf{R}_k$ be as in Proposition 3. It holds that

$$\begin{array}{l} \imath Out(\mathbf{P}_i) = \imath pr(B \wedge P[1] \wedge \cdots \wedge P[i]), \\ \imath Out(\mathbf{R}_k) = \imath pr(B \wedge P[1] \wedge \cdots \wedge P[k] \wedge Q[k]). \end{array}$$

*Proof*

We show the first clause by induction. The second clause is similar.

- Base case: $\mathbf{P}_0 = \imath FM(B, \emptyset) = (B^\omega,\ \emptyset,\ \imath pr(B))$.
- Inductive step: Assume that $\imath Out(\mathbf{P}_{i-1}) = \imath pr(B \wedge P[1] \wedge \cdots \wedge P[i-1])$. The module $\imath FM(P[i], \imath Out(\mathbf{P}_{i-1}))$ is

$$(P[i]^\omega,\ \imath Out(\mathbf{P}_{i-1}),\ \imath pr(P[i]) \setminus \imath Out(\mathbf{P}_{i-1}))\ .$$

Thus

$$\imath Out(\mathbf{P}_i) = \imath Out(\mathbf{P}_{i-1}) \cup \big(\imath pr(P[i]) \setminus \imath Out(\mathbf{P}_{i-1})\big) = \imath Out(\mathbf{P}_{i-1}) \cup \imath pr(P[i])$$

and by the I.H., this is then $\imath pr(B \wedge P[1] \wedge \cdots \wedge P[i])$.

∎

*Lemma 3*

Given any two first-order formulas $F_1, F_2$ and disjoint sets of predicate constants $\mathbf{p}_1, \mathbf{p}_2$ such that $\imath pr(F_1) \subseteq \mathbf{p}_1$, and $F_2$ is negative on $\mathbf{p}_1$. Every strongly connected component of $\mathrm{DG}[F_1 \wedge F_2; \mathbf{p}_1\mathbf{p}_2]$ is contained in $\mathbf{p}_1$ or $\mathbf{p}_2$.

*Proof*
Since $F_2$ is negative on $\mathbf{p}_1$, we have that $\imath head(F_2) \cap \mathbf{p}_1 = \emptyset$. Thus every outgoing edge in the dependency graph from a predicate constant in $\mathbf{p}_1$ must be obtained from $F_1$. Since $\imath pr(F_1) \subseteq \mathbf{p}_1$, such outgoing edge always leads to a vertex in $\mathbf{p}_1$. Consequently, every strongly connected component of $\mathrm{DG}[F_1 \wedge F_2; \mathbf{p}_1\mathbf{p}_2]$ containing a predicate constant from $\imath head(F_1)$ is contained in $\mathbf{p}_1$, so the claim follows. ∎

*Proposition 3*
If an incremental first-order theory $\langle B, P[t], Q[t]\rangle$ is acyclic, then the following modules are defined for all $k \geq 0$.

$$\begin{aligned}
\mathbf{P}_0 &= \imath FM(B, \emptyset),\\
\mathbf{P}_i &= \mathbf{P}_{i-1} \sqcup \imath FM(P[i], \imath Out(\mathbf{P}_{i-1})), && (1 \leq i \leq k)\\
\mathbf{R}_k &= \mathbf{P}_k \sqcup \imath FM(Q[k], \imath Out(\mathbf{P}_k))\ .
\end{aligned}$$

*Proof*
We first prove by induction that $\mathbf{P}_i$ is defined.

**Base case:** It is clear that $\mathbf{P}_0 = \imath FM(B, \emptyset)$ is defined.

**Inductive step:** Assume that $\mathbf{P}_{i-1} = (F_{i-1}, \mathcal{I}_{i-1}, \mathcal{O}_{i-1})$ is defined for any $i > 0$. Also,

$$\imath FM(P[i], \mathcal{O}_{i-1}) = (P[i]^\omega, \mathcal{O}_{i-1}, \imath pr(P[i]) \backslash \mathcal{O}_{i-1})$$

is trivially defined. To show that they are joinable, we will check the following:

(i) $\imath head(F_{i-1}) \cap (\imath pr(P[i]) \backslash \mathcal{O}_{i-1}) = \emptyset$;
(ii) $\imath head(P[i]^\omega) \cap \mathcal{O}_{i-1} = \emptyset$;
(iii) every strongly connected component of

$$\mathrm{DG}[F_{i-1} \wedge P[i]^\omega;\ \mathcal{O}_{i-1} \cup (\imath pr(P[i]) \backslash \mathcal{O}_{i-1})]$$

is a subset of $\mathcal{O}_{i-1}$ or $\imath pr(P[i]) \backslash \mathcal{O}_{i-1}$.

Note that

$$\imath pr(F_{i-1}) \subseteq \imath pr(B \wedge P[1] \wedge \dots P[i-1]) \tag{A3}$$

and

$$\imath head(P[i]^\omega) \subseteq \imath head(P[i])\ . \tag{A4}$$

*Proof of Claim (i):* By Lemma 2, $\mathcal{O}_{i-1}$ is $\imath pr(B \wedge P[1] \wedge \cdots \wedge P[i-1])$, and Claim (i) trivially follows in view of (A3) and the fact that $\imath head(F_{i-1}) \subseteq \imath pr(F_{i-1})$.

*Proof of Claim (ii):* Since the theory is acyclic,

$$\imath head(P[i]) \cap \imath pr(B \wedge P[1] \wedge \cdots \wedge P[i-1]) = \emptyset\ ,$$

and from (A4) and Lemma 2, we have that

$$\imath head(P[i]^\omega) \cap \mathcal{O}_{i-1} = \emptyset\ . \tag{A5}$$

*Proof of Claim (iii):* The claim follows from (A5) and Lemma 3.

We next show that $\mathbf{R}_k$ is defined. By our previous result, $\mathbf{P}_k = (F_k, \mathcal{I}_k, \emptyset_k)$ is defined. It also holds that

$$\imath FM(Q[k], \emptyset_k) = (Q[k]^\omega, \emptyset_k, \imath pr(Q[k]) \setminus \emptyset_k)$$

is defined trivially. The rest of the reasoning is similar to the previous one.

### *A.7 Proof of Proposition 4*

*Proposition 4*
Let $\langle B, P[t], Q[t]\rangle$ be an acyclic incremental first-order theory and let $\mathbf{R}_k$ be the module as defined in the statement of Proposition 3. For any nonnegative integer $k$,

$$\begin{aligned} I_B \cup I_{P[1]} \cup \cdots \cup I_{P[k]} \cup I_{Q[k]} \models \mathrm{SM}[\mathbf{R}_k] & \\ \text{iff} \quad I_B \models \mathrm{SM}[\imath FM(B, \emptyset)] & \\ \text{and} \quad I_{P[1]} \models \mathrm{SM}[\imath FM(P[1], \imath Out(\mathbf{P}_0))] & \\ \text{and} \quad \ldots & \\ \text{and} \quad I_{P[k]} \models \mathrm{SM}[\imath FM(P[k], \imath Out(\mathbf{P}_{k-1}))] & \\ \text{and} \quad I_{Q[k]} \models \mathrm{SM}[\imath FM(Q[k], \imath Out(\mathbf{P}_k))] & \ . \end{aligned}$$

where $I_B$ ($I_{P[1]}, \ldots, I_{P[k]}, I_{Q[k]}$, respectively) is a $\mathbf{c}(B)$-partial interpretation ($\mathbf{c}(P[1])$, $\ldots, \mathbf{c}(P[k]), \mathbf{c}(Q[k])$-partial interpretation, respectively) such that $I_B, I_{P[1]}, \ldots, I_{P[k]}, I_{Q[k]}$ are pairwise compatible.

*Proof*
Via repeated applications of Theorem 4 on $\mathbf{R}_k$ as indicated by Proposition 3. ∎

### *A.8 Proof of Proposition 5*

*Lemma 4*
Let $\langle B, P[t], Q[t]\rangle$ be an acyclic incremental first-order theory, let $k$ be a nonnegative integer, let $H_k = B \wedge P[1] \wedge \cdots \wedge P[k]$, and let $R_k$ be the $k$-expansion of the incremental theory. It holds that $I_B \cup I_{P[1]} \cup \cdots \cup I_{P[k]} \cup I_{Q[k]} \models \mathrm{SM}[R_k]$ iff

$$\begin{aligned} & I_B \models \mathrm{SM}[B;\ \imath pr(B)] \\ & \quad \text{and} \quad I_{P[1]} \models \mathrm{SM}[P[1];\ \imath pr(P[1]) \setminus \imath pr(H_0)] \\ & \quad \text{and} \quad \ldots \\ & \quad \text{and} \quad I_{P[k]} \models \mathrm{SM}[P[k];\ \imath pr(P[k]) \setminus \imath pr(H_{k-1})] \\ & \quad \text{and} \quad I_{Q[k]} \models \mathrm{SM}[Q[k];\ \imath pr(Q[k]) \setminus \imath pr(H_k)] \end{aligned} \qquad \text{(A6)}$$

where $I_B$ ($I_{P[1]}, \ldots, I_{P[k]}, I_{Q[k]}$, respectively) is a $\mathbf{c}(B)$-partial interpretation ($\mathbf{c}(P[1])$, $\ldots, \mathbf{c}(P[k]), \mathbf{c}(Q[k])$-partial interpretation, respectively) such that $I_B, I_{P[1]}, \ldots, I_{P[k]}, I_{Q[k]}$ are pairwise compatible.

*Proof*
Formula $H_k$ is trivially negative on $\imath pr(Q[k]) \setminus \imath pr(H_k)$, and since the theory is acyclic, $Q[k]$ is negative on $\imath pr(H_k)$. Also, by Lemma 3, every strongly connected component of $\mathrm{DG}[H_k \wedge Q[k];\ \imath pr(H_k) \cup \imath pr(Q[k])]$ is a subset of $\imath pr(H_k)$ or $\imath pr(Q[k]) \setminus \imath pr(H_k)$. By Theorem 4, it then holds that

$$I_{H_k} \cup I_{Q[k]} \models \mathrm{SM}[R_k] \quad \text{iff} \quad I_{H_k} \models \mathrm{SM}[H_k] \text{ and } I_{Q[k]} \models \mathrm{SM}[Q[k];\ \imath pr(Q[k]) \setminus \imath pr(H_k)]$$

where $I_{H_k}$ is a $\mathbf{c}(H_k)$-partial interpretation that is compatible with $I_{Q[k]}$.

Next we check by induction that $I_{H_k} \models \mathrm{SM}[H_k]$ is equivalent to

$$\begin{array}{l} I_B \models \mathrm{SM}[B] \\ \quad \text{and } \ I_{P[1]} \models \mathrm{SM}[P[1];\ \imath pr(P[1]) \setminus \imath pr(H_0)] \\ \quad \text{and } \ \dots \\ \quad \text{and } \ I_{P[k]} \models \mathrm{SM}[P[k];\ \imath pr(P[k]) \setminus \imath pr(H_{k-1})]\ . \end{array} \tag{A7}$$

**Base case:** when $k = 0$, $H_k = B$. Trivial.

**Inductive step:** Let the property hold for $H_{k-1}$. By definition, $H_k = H_{k-1} \wedge P[k]$. $H_{k-1}$ is trivially negative on $\imath pr(P[k]) \setminus \imath pr(H_{k-1})$ and since the theory is acyclic, $P[k]$ is negative on $\imath pr(H_{k-1})$. Also, by Lemma 3, every strongly connected component of $\mathrm{DG}[H_k; \imath pr(H_k)]$ is a subset of $\imath pr(H_{k-1})$ or $\imath pr(P[k]) \setminus \imath pr(H_{k-1})$. By Theorem 4, it then holds that

$$I_{H_k} \models \mathrm{SM}[H_k] \quad \text{iff} \quad I_{H_{k-1}} \models \mathrm{SM}[H_{k-1}] \text{ and } I_{P[k]} \models \mathrm{SM}[P[k];\ \imath pr(P[k]) \setminus \imath pr(H_{k-1})].$$

The property then holds by the I.H. ∎

*Lemma 5*
For any first-order formula $F$, $\mathrm{SM}[\imath FM(F, \mathcal{I})]$ is equivalent to $\mathrm{SM}[F; \imath pr(F) \setminus \mathcal{I}]$.

*Proof*
We introduce a notion that helps us prove. By $\imath Simpl(F)$ we denote the least fixpoint of the sequence $F_0, F_1, \dots$: formula $F_0$ is defined as $F$, and $F_{i+1}$ is defined as $F_i|_{head(F_i)}$.

Formula $\mathrm{SM}[\imath FM(F, \mathcal{I})]$ is $\mathrm{SM}[(F^\omega, \mathcal{I}, \imath pr(F) \setminus \mathcal{I})]$, which in turn is defined as $\mathrm{SM}[F^\omega;\ \imath pr(F) \setminus \mathcal{I}]$. By Theorem 2 from (Ferraris et al. 2011), this is equivalent to $\mathrm{SM}[F^\omega \wedge \imath Choice(\mathcal{I});\ \imath pr(F)]$. From the definition of $\imath Simpl$, the latter is equivalent to $\mathrm{SM}[Simpl(F \wedge \imath Choice(\mathcal{I}));\ \imath pr(F)]$, and, furthermore, by Theorem 4 from (Ferraris et al. 2011), is equivalent to $\mathrm{SM}[F \wedge \imath Choice(\mathcal{I});\ \imath pr(F)]$. ∎

*Proposition 5*
Let $\langle B, P[t], Q[t] \rangle$ be an acyclic incremental theory, let $k$ be a nonnegative integer, let $R_k$ be the $k$-expansion of the incremental theory, and let $\mathbf{R}_k$ be the module as defined in Proposition 3. For any $\mathbf{c}$-partial interpretation $I$ such that $\mathbf{c} \supseteq \mathbf{c}(R_k)$, we have that

$$I \models \mathrm{SM}[R_k] \ \text{ iff } \ I \models \mathrm{SM}[\mathbf{R}_k].$$

*Proof*

Without loss of generality, let $I = I_B \cup I_{P[1]} \cup \cdots \cup I_{P[k]} \cup I_{Q[k]}$. By Lemma 4, $I \models \text{SM}[R_k]$ is equivalent to (A6), and by Lemma 2, this is further equivalent to

$$\begin{aligned} & I_B \models \text{SM}[B;\ \imath pr(B)] \\ & \quad \text{and } \ I_{P[1]} \models \text{SM}[P[1];\ \imath pr(P[1]) \setminus \imath Out(\mathbf{P}_0)] \\ & \quad \text{and } \ \ldots \\ & \quad \text{and } \ I_{P[k]} \models \text{SM}[P[k];\ \imath pr(P[k]) \setminus \imath Out(\mathbf{P}_{k-1})] \\ & \quad \text{and } \ I_{Q[k]} \models \text{SM}[Q[k];\ \imath pr(Q[k]) \setminus \imath Out(\mathbf{P}_k)]\ . \end{aligned}$$

We check the following:

- $I_B \models \text{SM}[B;\ \imath pr(B)]$ iff $I_B \models \text{SM}[\imath FM(B, \emptyset)]$;
- $I_{P[i]} \models \text{SM}[P[i];\ \imath pr(P[i]) \setminus \imath Out(\mathbf{P}_{i-1})]$ iff $I_{P[i]} \models \text{SM}[\imath FM(P[i], \imath Out(\mathbf{P}_{i-1}))]$;
- $I_{Q[k]} \models \text{SM}[Q[k];\ \imath pr(Q[k]) \setminus \imath Out(\mathbf{P}_k)]$ iff $I_{Q[k]} \models \text{SM}[\imath FM(Q[k], \imath Out(\mathbf{P}_k))]$.

The first clause is clear. The last two clauses follow from Lemma 5.

Therefore, by Proposition 4,

$$I_B \cup I_{P[1]} \cup \cdots \cup I_{P[k]} \cup I_{Q[k]} \models \text{SM}[\mathbf{R}_k]\ .$$

∎